# A Control Strategy for an Autonomous Robotic Vacuum Cleaner for Solar Panels

Aravind G, Gautham Vasan*, Gowtham Kumar T.S.B, Naresh Balaji R
G. Saravana Ilango
National Institute of Technology - Tiruchirapalli
Tiruchirapalli - 620015
Email: *gauthamv.529@gmail.com

*Abstract—* Accumulation of dust on the surface of solar panels reduces the amount of radiation reaching it. This leads to loss in generated electric power and formation of hotspots which would permanently damage the solar panel. This project aims at developing an autonomous vacuum cleaning method which can be used on a regular basis to maximize the lifetime and efficiency of a solar panel. This system is implemented using two subsystems namely a Robotic Vacuum Cleaner and a Docking Station. The Robotic Vacuum Cleaner uses a two stage cleaning process to remove the dust from the solar panel. It is designed to work on inclined and slippery surfaces. A control strategy is formulated to navigate the robot in the required path using an appropriate feedback mechanism. The battery voltage of the robot is determined periodically and if it goes below a threshold, it returns to the docking station and charges itself automatically using power drawn from the solar panels. The operation of the robotic vacuum cleaner has been verified and relevant results are presented. The DC Charging circuit in the docking station is simulated in Proteus environment and is implemented in hardware. An economical, robust Robotic Vacuum Cleaner which can clean arrays of Solar panels (with or without inclination) interlinked by rails and recharge itself automatically at a docking station is designed and implemented.

## I. INTRODUCTION

India is one of the few countries endowed with abundant solar energy. The country receives about 5000Trillion KWh/year, which is more than sufficient to satisfy the power requirement of the entire nation [1]. With more than 300 million people without access to uninterrupted supply of electricity and industries citing energy shortage as key growth barrier in India, solar power has the potential to help the country address the shortage of power for economic growth [2].

Although solar energy is a very promising source since it supports delocalized power generation, it requires regular maintenance after installation. Accumulated dust on the surface of photovoltaic solar panel can reduce the system's efficiency up to 50% [3] [4]. This emphasizes the need to keep the surface of the solar panel as clean as possible. Most of the present cleaning methods employ water based techniques (e.g., washing directly from a water pump, a soap solution, etc.). One could not afford to waste copious amounts of water on cleaning solar panels since it cannot be recycled very easily for practical uses. Also standalone panels installed in different areas do not always have a nearby water source which further adds to the problem. Another efficient method of cleaning solar panels is using Electrostatic cleaning method. Though it is very effective, there is a decrease in power performance [5]. Also every solar panel requires an individual electrostatic cleaner for permanent installations which makes cleaning laborious and expensive [5].

Photovoltaic panels are generally installed in relatively inaccessible areas like roofs or arid deserts which make manual cleaning operations difficult and expensive. Most solar panels are normally cleaned early in the morning or late at night since cleaning during its principal operation leads to non-uniform power outage and decrease in efficiency. Thus the lack of automation capabilities in most cleaning solutions proves costlier in terms of water and energy-use. Thus, by implementing the proposed design the need for water based cleaning methods, manual intervention and cleansing difficulties in remote places is eliminated.

The goal of the project is to design a robust, commercially viable product which provides a simple, cost-effective solution to the clean solar panels. The robot uses a two stage cleaning process to remove dust effectively from the solar panels. A rolling brush is placed in front to disperse the dust towards the vacuum cleaner. A high speed motor capable of creating suitable suction is used for removing dust from the panels. It traverses the solar panel using a pre-defined path controlled by the accelerometers and ultrasonic sensor. The proposed design can detect edges easily, work on inclined planes and automatically charge itself at the docking station.

The report has been organized into 6 sections as follows: section 1 provides the comparison with previous works, section 2 gives the technical background for the project, section 3.1 and 3.2 provide the detailed explanation of the proposed system, section 4.A shows the practical realization of the system in hardware with different ICs, and section 4.B presents the implementation of the various algorithms using flowcharts. Section 5 shows the various experimental results and simulations performed to verify the system. The conclusions and future scope of the project are presented in Section 6

## II TECHNICAL BACKGROUND

When there is a loss of illumination intensity caused by obscuration of light by dust layers on the panels, there are three adverse effects: (1) Reduction in power output (2) Decrease in overall efficiency (3) Formation of hotspots and dead cells if the modules are partially blocked by dust layer deposits [6]. When some of the cells are covered by dust, the shaded cells do not generate enough power to match the other cells; rather, they act as dead load on the working cells. As a result, the temperature of the shaded cell increases forming hotspots [6]. Unless efficient protection devices like bypass diodes [7] are used to prevent the formation of hotspots, the modules can get permanently damaged [6].

Hence the need for a cost-effective, automated cleaning solution arises which can be satisfied by the proposed method. Though the concept of a robotic vacuum cleaner has already been commercially implemented by corporations like Electrolux (Trilobite), iRobot (Roomba), the problem lies in the fact that they are designed for home applications only. When the concept of a robotic vacuum cleaner for solar panels comes into picture, the factors to be taken in to account are: (1) Inclination of the Solar Panels (2) Multiplicity of panels (3) Detection of edges (4) Scheduled time of operation (5) Need for a generic algorithm.

Though it can be argued that vacuum cleaner designed for home applications like Roomba can be used on solar panels, there are few complications that arise as follows: The area of cleaning is fixed by either allowing Roomba to find the perimeter using its IR sensor or by manually setting up to a maximum of two virtual walls [8]. But solar panels do not have any large obstacles (vertical structures/walls) in the surrounding area and hence to find the perimeter, it now has to use four virtual walls which is not possible. To avoid this, a set of rules are pre-defined so that the robot traverses along a pattern until it reaches the end of the interconnected array of panels. Also Roomba, Trilobite, etc., are not designed specifically for working on inclined planes. Hence, problems like slipping, power wastage, etc., may arise. This difficulty is tackled by using powerful Geared DC motors coupled with gripper wheels. Also Roomba finds the shortest path back to the docking station from its virtual map in the memory and a RF module [9]. But since it is very unlikely to create a memory map, it may not reach the docking station via a computed route, but through a more complex route. A simple yet effective path is used in the proposed system. Roomba employs Infra-Red sensors for obstacle detection, perimeter calculation and cliff detection [8]. It also uses the Infra-Red signal to find its current location in the room [10]. But IR sensors are not very effective when used in sunlight. Instead an Ultrasonic sensor has been placed in the front of the robot to detect edges of the panels/cliffs and stop in order to change the direction of motion. The robot's position in the panel is found by counting the number of cycles of cleaning.

To cover the entire expanse of the solar panels, a path is pre-defined as shown in fig 2.1. The robot travels in a reciprocating motion to travel along the defined path. Feedback from ultrasonic sensor and accelerometers are used to detect the edges (cliffs) and stay on the defined path. The arrays of solar panels are interlinked using rails to ensure that one robot is enough to clean an entire array of solar panels. Once a group of solar panels are cleaned completely, the robot moves to the next group using the connecting rails. For the robot to be fully autonomous, it should be able to dock itself and charge itself automatically. Thus when the battery level is low, the robot stops the cleaning process and returns to the docking station to charge itself. Provision of battery charging is provided in the docking station by using a Lithium Polymer battery charging circuit. This circuit draws input power supply from the solar panels. When charging is complete, the robot disconnects itself from the docking station. Now the robot starts cleaning the panels if it is required to do so. The Overall Block Diagram of this system is shown in fig 2.2.

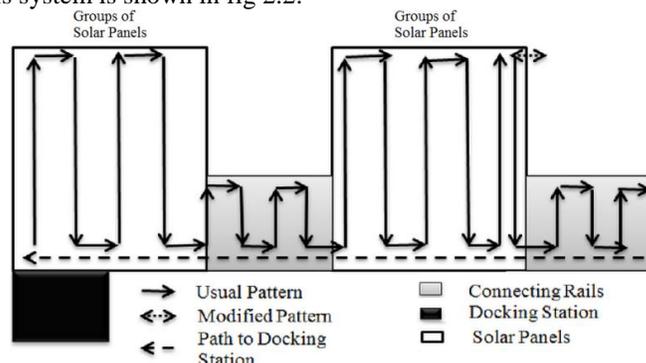

**Fig 2.1: Path-Plan Overview**

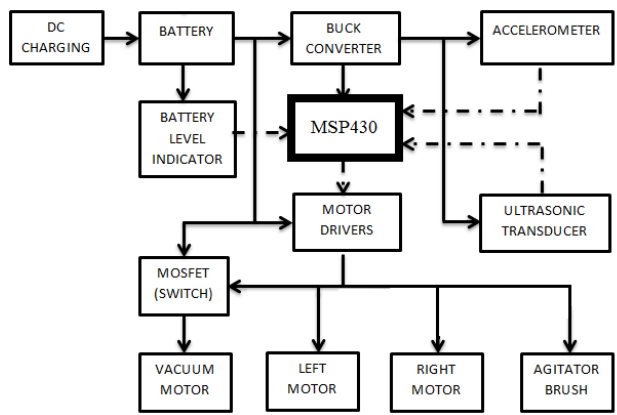

**Fig 2.2: Overall Block Diagram**

## III. PROPOSED SYSTEM

The proposed system is implemented by developing two subsystems as follows:

**(i) The Robotic Vacuum Cleaner**
The dust accumulated on solar panels forms a sticky layer which cannot be cleaned directly by using portable vacuum cleaners. Hence a two stage cleaning process is implemented. *Stage 1:* A rolling brush is fixed on the robot such that it agitates and pushes the dust towards the vacuum cleaner. *Stage 2:* The vacuum motor is used to create enough suction to collect the dust scattered on the solar panel. The presence of a sticky layer of dust on a smooth inclined surface adds to the problem of slipping. Therefore, to have better traction gripper wheels are used to traverse the solar panels. The robot is controlled using the MSP430G2553 microcontroller. It acts as the master control element of the robot. The robot is designed to minimize the total load in order to achieve higher efficiency and longer battery life.

**(ii) The Docking Station**
The docking station is setup at the beginning of the solar panels. It comprises of a base and two Aluminium strips mounted on it acting as the positive and negative terminals. The circular design ensures that the robot can charge itself at any orientation till it makes contact with the charging strips.

If the battery voltage falls below a particular threshold, the robot returns to the docking station to charge itself. During the charging process, if the battery voltage exceeds the reference limit, the charging circuit disconnects itself from the battery. When charging is complete, it starts the cleaning the panels again if required.

### 3.1 Hardware Components
#### 3.1.1 MSP430 Launchpad
The MSP430G2 Launchpad is used for the overall control of the robot. It triggers the ultrasonic sensor and processes the ADC inputs from the accelerometer and echo pulses from the ultrasonic sensor for navigation control. It also relays control signals to the Motor driver. It takes input signals from the battery level indicator to keep track of the source battery voltage. A common output signal from the MSP430 is used to control the rolling brush and the vacuum cleaner simultaneously. Since the vacuum cleaner requires a large amount of current, it is connected directly to the battery and controlled using a Power-MOSFET IRFZ44 as a switch.

#### 3.1.2 Lithium Polymer Battery Charging Circuit
The DC charging circuit comprises of a Lithium Battery charging circuit which is placed in the docking station. It is a Six Cell 12.6V charger with a DC input of 15V – 40V. The variable voltage regulator LM317 transistor BC547, dual operational amplifier LM358N and Power-MOSFET IRF540N are used in the circuit. An over-voltage protection circuit is added to the charging circuit to ensure safety. The rate at which the battery continues to absorb charge or the current from the solar panels gradually slows down because the voltage is maintained constant. Completion of charging is indicated by the use of an LED.

#### 3.1.2 DC-DC Converter
The buck converters LM2675-5.0 and LM2675-3.3 are used to regulate the power supply. It also comprises of a diode and LC filter. The LC filter is used to achieve a ripple free DC output.

#### 3.1.3 Battery Level Indicator
LM 3914 (Dot/Bar Led Driver) has been used to indicate the voltage of the battery. It is operated in dot mode. It is calibrated for the full charge voltage (12.6 V) of the battery.

#### 3.1.4 Motor Driver
The TPIC0298 motor driver module has been used to control the motors since they require a higher current which cannot be drawn using the L293D motor driver. The TPIC0298 can provide a maximum of 2A per channel.

#### 3.1.5 Navigation Control
In order to ensure the robot travels in the right path, two accelerometers and ultrasonic sensors are used to control the orientation and detect edges respectively. The MMA7361 Triple axis accelerometers are used with sensitivity 800 mV/g. The HCSR04 Ultrasonic sensor module is used to find cliffs/obstacles along the path.

### 3.2 Software Components
#### 3.2.1 Path-Planning Algorithm
The robot traverses the path as shown in fig 2.2. This pre-defined path ensures that the total expanse of the solar panels is covered effectively. The Ultrasonic sensor is used detect the edges (cliffs) of the solar panel. In order to stay on the defined path, the output signals from the accelerometers are processed by the MSP430G2553 Microcontroller and compared with a pre-defined set of values. The Proportional Integral Derivative (PID) Control technique [11] is implemented by adding the calculated error to the Timer's capture/compare register value which alters the Pulse Width Modulation (PWM) signal given as input to the motor driver. The total error, e is found by using the formula:

$$e = e_p \times k_p + e_i \times k_i + e_d \times k_d$$

Where $e_p$, $e_i$, $e_d$ are the proportional, integral and derivative errors found by comparing the accelerometer readings with preset values. $k_p$, $k_i$ and $k_d$ are found by repeated testing on inclined surfaces. A particular set of values are chosen such that they are applicable for any small inclination. The PID control algorithm is used only to control the navigation of the robot so that the total expanse of the solar panels is covered effectively.

### 3.2.2 Battery Charging Algorithm
In an attempt to improve the battery health, two states of charging have been implemented. The battery is charged at to a preset threshold of 12.6 V beyond which it is charged at a lower current to prevent damage.

### 3.3 Assumptions
- The inclination of the solar panels is limited to 30 degrees.
- Only dust particles or small dispersible impurities are settled on the solar panels. Water or any other substances which can cause slipping are not present on the surface.
- The dust is collected every week using the robot.
- Though rails are used and a docking station as a part of this project, it is generally not provided in regular arrangements. Hence the provision of rails and a docking station is absolutely necessary for that robot to clean all panels and recharge automatically.
- The differential turns taken by the robot are assumed to be zero radius turns.

### 3.4 Constraints
- When the robot moves on inclined surfaces, it can encounter a free fall due to gravitational force. Hence speed of the robot is regulated and gripper wheels are used.
- In most installations, a bump is formed at the point of contact between the two solar panels. The robot needs additional power to climb over this obstacle. Hence small gripper wheels are used instead of castor wheels to climb smoothly over the bump.
- Though negative power supply can be provided using a TI IC DN43, it has not been used along with IC741 in order to reduce the complexity and size of the charging circuit on the docking station. Hence LM358, a single-supply dual operational amplifier has been used for this purpose.

### IV IMPLEMENTATION

*A. Hardware Implementation*
The hardware components of the system include the microcontrollers, DC-DC converters, sensors, voltage regulators, battery level indicator and motor drivers. The Robot is self-sufficient since it can be powered by the PV panel array does not require any other external power source. The voltage sensors are constructed with differential amplifier circuits using appropriate op-amps. The microcontroller is powered from 3.3V supplies obtained via Buck converter (LM2675-3.3). The robotic vacuum cleaner has the following analog circuits onboard: (a) Master Control Board (b) Buck Converter (c) Battery Level Indicator (d) Motor Driver. The docking station houses the robot and the Lithium polymer battery charging circuit.

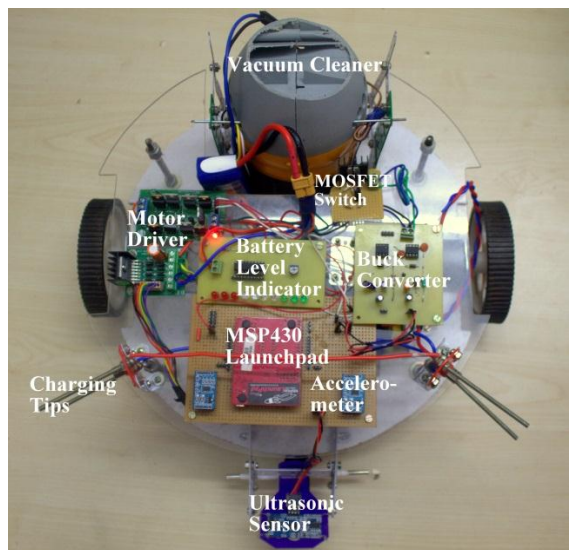
**Fig 4.1: Hardware Setup**

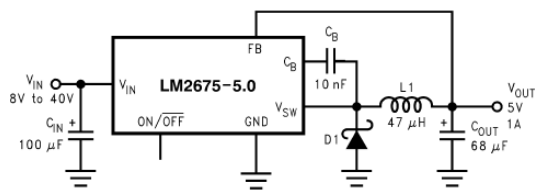
**Fig 4.2: Buck Converter**

### 4.1 Master Control Board
This board is a general purpose board (shown in fig 4.1 and 4.6) designed to connect the required pins from the Accelerometer and Ultrasonic sensor to the MSP430G2553 Launchpad. It was primarily designed keeping in mind the ease of usage and mounting difficulties of the Launchpad on the robot. It also relays control signals to the Motor driver. The output voltages of the two buck converter power the board. The battery level indicator is also connected to the Microcontroller so that the battery voltage is continuously monitored.

### 4.2 Buck Converter
The buck converter module (fig.4.2) consists of two individual buck converter ICs LM2675-3.3V and LM2675-5.0V. It also comprises of a diode and LC filter. The LC filter is used to achieve a ripple free DC output.

The values of L and C were found using the formula:
$$L = \frac{((V_{pv} - V_{battery}) \times D)}{(f \times 2\Delta i)}$$
And
$$f = \frac{1}{2\pi\sqrt{LC}}$$

Though a regular variable voltage regulator would suffice for the power requirements, an efficient solution is needed to guarantee a longer battery life. The vacuum cleaner alone draws a lot of power. Added to this power is required for running three other motors and overcome the forces of gravity and friction. Hence an energy efficient alternative is required.

### 4.3 Battery Level Indicator
The LM3914 dot/bar display driver is used to indicate the voltage level based on battery voltage (as shown in fig 4.3). It is operated in dot mode and calibrated for a voltage of 12.6V using appropriate resistances.

### 4.4 Motor Driver
The TPIC0298 motor driver (fig.4.4) is used for controlling the motors as necessary. It is capable of supplying up to 2A per channel. The circuit also uses 8 fly-back diodes protect the IC from back-EMF. It is capable of controlling two DC motors. The vacuum cleaner draws a large amount of current and hence can damage the motor driver. Hence the control signal given to the rolling brush is also given to a Power-MOSFET which accordingly switches on/off the supply to the vacuum cleaner.

### 4.5 Lithium Polymer Battery Charging Circuit
In the DC charging circuit (fig.3.4), a variable voltage regulator LM317 is used to set the required maximum charge voltage at 12.6V. In this circuit, resistor R4 is used to effectively limit the current output. When the voltage across R4 reaches the threshold of +0.7V at the base of the transistor, the transistor BC547 starts to conduct and brings the ADJ pin progressively to ground. Heat sinks are used along with the LM317 since it heats up a lot due to the high power dissipation. Once charging is complete, the op-amp switches off the transistor and hence charging is stopped and hence the LED stops glowing.

It is absolutely necessary to immediately disconnect the battery from the supply when it is fully charged. Otherwise, it is very likely that the battery will explode. Hence a protection circuit is added by using LM358N, a single-supply dual operational amplifier and a Power-MOSFET IRF540N. The power-MOSFET is used since it has a higher current rating and satisfies the requirements. The MOSFET is used as a switch in this circuit. The Op-amp is used as a comparator. As the battery charges, voltage across resistor R4 varies. This voltage is compared with the reference voltage. The reference voltage is generated using a sensing diode 1N4148 and a potential divider circuit through the LM358N. When voltage across R4 exceeds the threshold the Op-Amp goes low and hence switches off the MOSFET which disconnects the battery from charging.

### 4.6 Hardware
The robot is built on a circular wooden chassis. It is powered by two DC geared motors placed diametrically opposite to each other. The wheels used are made of plastic and they have an additional gripper made of rubber. It has a diameter of 11.3 cm which provides ample grip and stability. On the bisecting diameter two small rubber wheels are placed. The robot uses a differential drive for steering. The usage of differential drive permits zero radius turns and easy steering. The motors used are 12V, 5 Kg-cm torque, 60RPM Side-Shaft Geared DC motors. The rolling brush is coupled with a gear arrangement shown in fig.4.7. The gears are attached to a 12V, 300RPM Side-Shaft DC Geared motor. The vacuum motor is a 20000 RPM Johnson motor which has a power rating of 250W. A slot is created on the chassis where the vacuum cleaner is fixed such that it is placed just above the ground. The vacuum cleaner is held steady by using mechanical supports. Its base design is built using a customized carpet cleaner body and a filter. All the circuits are mounted on an acrylic sheet balanced using studs on the wooden chassis. The ultrasonic sensor is mounted on the front. The overall design is shown in fig 4.6. The rails and docking station (fig 4.7) are constructed using wood plank of appropriate dimensions. They're attached to the solar panel and the ground for sturdy mechanical support.

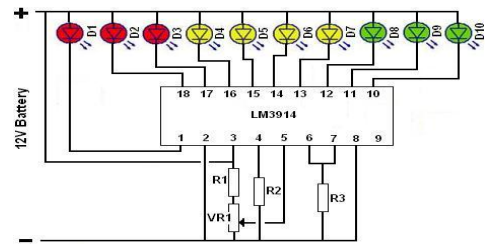
**Fig 4.3: 12V Battery Level Indicator**

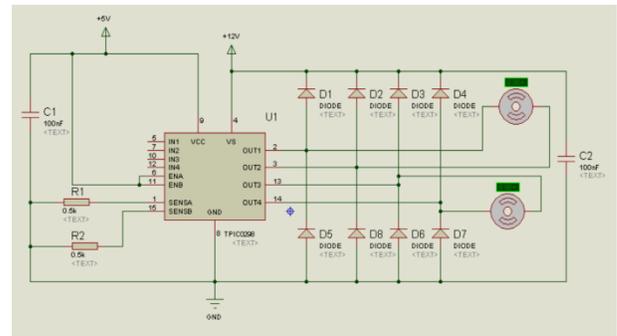
**Fig 4.4: Motor Driver Circuit**

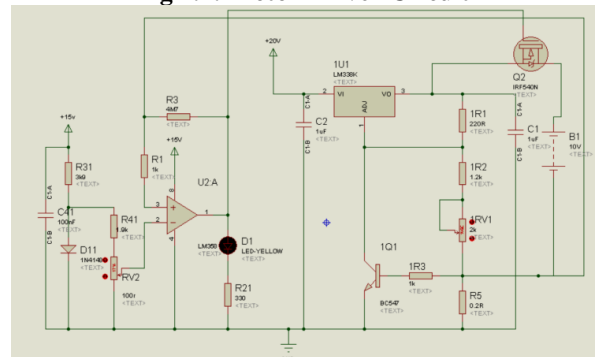
**Fig 4.5: DC Lithium Polymer Battery Charging Circuit**

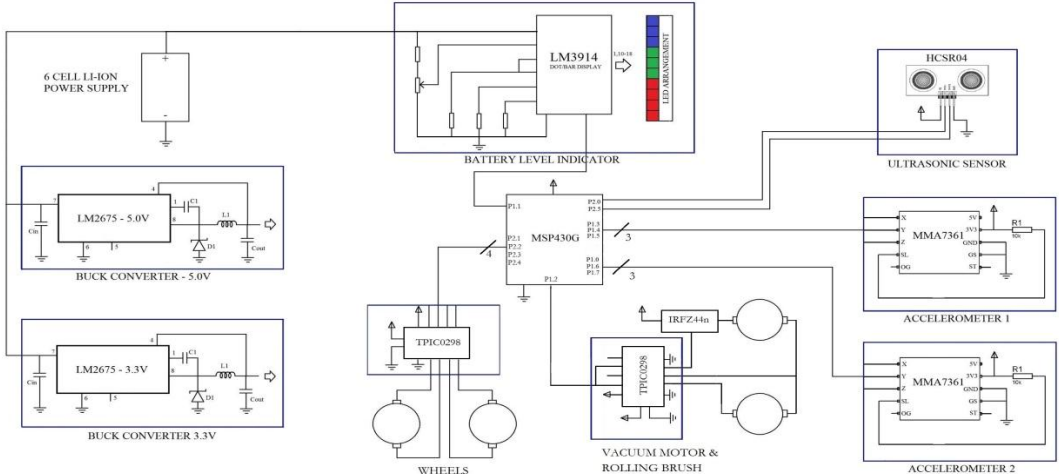
Fig 4.6: Overall Circuit Diagram

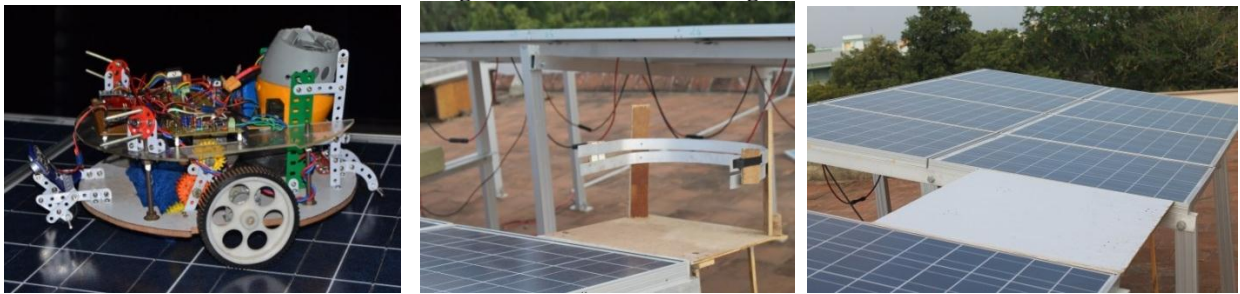
Fig 4.7: (i) Robotic Vacuum Cleaner (ii) Docking Station (iii) Connecting Rails

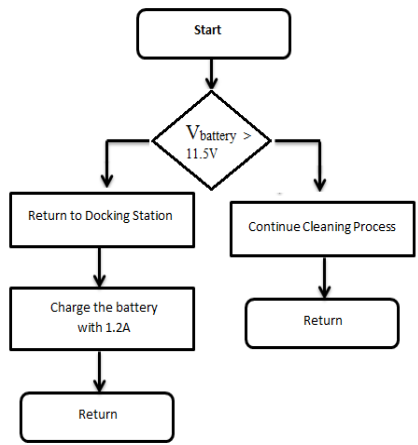
Fig 4.8: Battery Charging Algorithm

### B. Software Implementation

The complete algorithm for path-planning, PID control and battery charging is implemented using the MSP430G2553 microcontroller. The code was primarily tested using Code Composer Studio which interfaces the microcontroller using the JTAG interface. The data from various sensors were analyzed by sending their output data to the Computer using UART serial communication. The overall path planning algorithm is implemented as shown in fig 4.9. It uses ADC inputs from accelerometer and compares it with a preset value. This is to ensure that the robot exhibits linear motion.

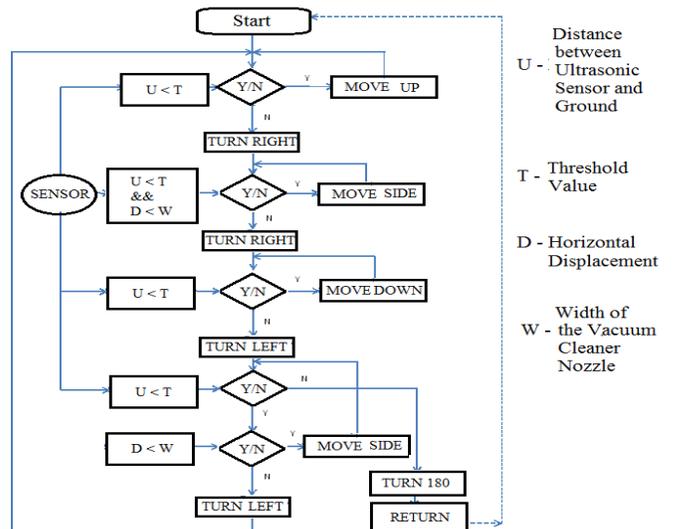
Fig 4.9: Path Planning Algorithm

The MSP430 triggers the ultrasonic sensor periodically and continuously calculates the distance between the robot and the ground using the echo pulses. This distance is also compared to a threshold which enables the robot to act accordingly whenever it encounters a cliff. The flowchart illustrated in fig 4.8 is used to check if the robot can continue cleaning the solar panels with its present battery state. If the voltage goes below a minimum threshold level, the robot completes the current cycle and it returns to the docking station. Once charging is complete, it resumes cleaning at the same column where it had left off earlier.

## V. RESULTS

The real-time navigation of the robot on the solar panel is verified by plotting the distance measured from ultrasonic sensor and input to both motors against time (shown in fig 5.1). When the system starts its cleaning process by moving out of the docking station, the distance computed from the readings of ultrasonic sensor is around two inches. This is the distance from the sensor to the panel surface below and is considered safe for the robot to move forward. The robot moves up the panel (a) (marked in the graph 5.1), correcting its orientation by tuning to the preset values of accelerometer until the edge of the panel, which is detected by a sudden surge in distance computed from the ultrasonic sensor. Now, the robot makes a differential turn (b) to follow the pre-computed path, which is seen by reversal of input voltage polarity to the right motor. When the 90 degree turn is completed, the robot moves a short distance along the panel longitudinally, a distance equal to the nozzle width of the vacuum cleaner. This ensures the robot successfully traverses along the panel without repeatedly moving over a single area. Now, the robot moves down the panel (c) with a reduced input to the motor compared to upward motion. This is to ensure constant velocity, taking into account the acceleration due to gravity. Again a differential turn is made, seen by the reversal of voltage to the left motor (d). This cycle is repeatedly implemented until the entire solar panel is cleaned completely.

The plot for PID control mechanism is shown in fig 5.2. The sudden change in the accelerometer readings arises when the robot tries to move over a bump at the point of contact of the solar panels. Initially the robot moves up correcting it orientation by tuning to the preset values (A). It can be seen that the speed reduces drastically when it tries to climb over the obstacle. In the beginning, the left motor goes over the bump. The robot tries to stay on the right orientation and hence reduces the speed of the left motor and increases the speed of the right motor (B) (marked in the graph 5.2). Once the robot reaches the correct orientation, the same low speed signal is given till it climbs down the bump (C). Immediately after it crosses the bump, the speed of the motors increases its speed and tunes itself automatically (D). Other disturbances are corrected automatically by sending signal to the motors.

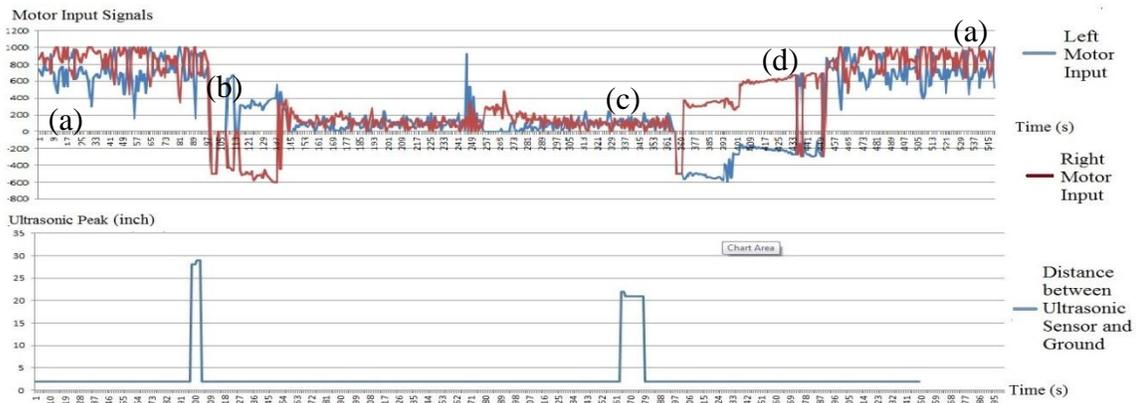
Fig 5.1: Ultrasonic sensor & Motor Input Signals Vs Time

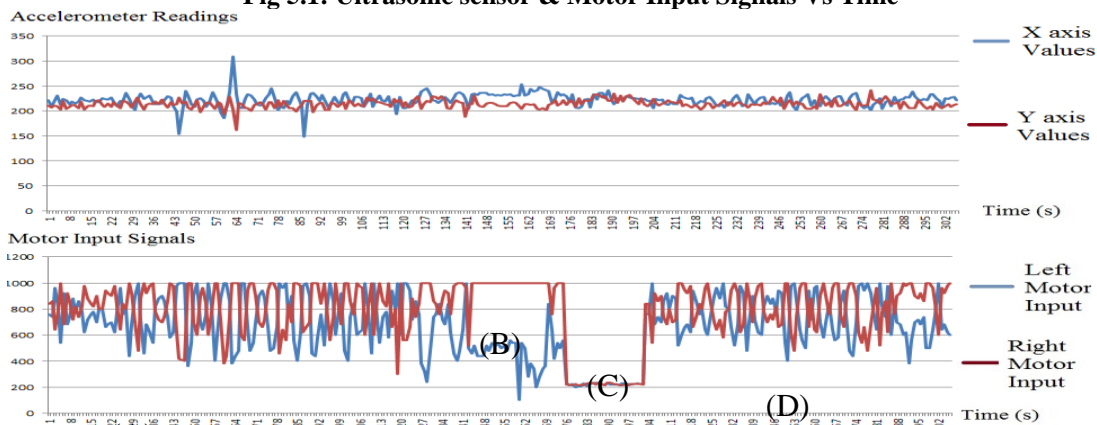
Fig 5.2: PID Control – Accelerometer outputs & Motor Inputs Vs Time

The fig 5.3 shows the result of the cleaning mechanism. The fig 5.4 shows the result of cleaning only once by using the robot to clean only the solar panel located in the Centre.

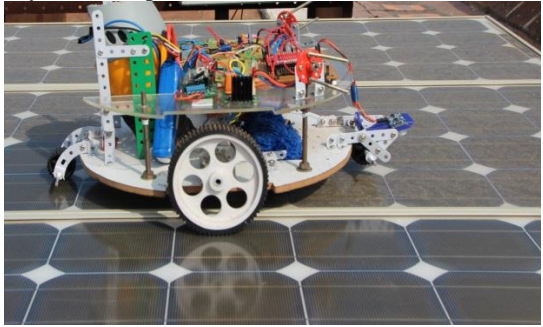

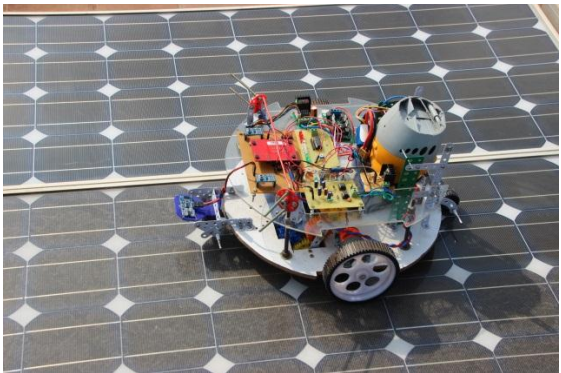

**Fig 5.3: Cleaning Process**

If the robot battery voltage falls below a particular threshold, The complete cleaning mechanism is shown in [12]. The overall working principle and circuit design are demonstrated in [12].

## VI. CONCLUSIONS

A control strategy for the Robotic Vacuum Cleaner has been designed and tested. The two stage cleaning mechanism is very effective in removing dust settled for a long time on solar panels. Cleaning two to three times instead of just once shows an improved performance in terms of effectiveness of cleaning. It has been proved experimentally that the robot can traverse on inclined surfaces without any difficulty. The robot traverses entire arrays of the solar panels interconnected using rails completely and clean with satisfactory results. It travels at a speed of '2-6' cm per second based on the inclination of the solar panel. If battery level falls below a threshold, it is confirmed that the robot returns to the docking station and charges itself automatically. It is verified that the robot returns to the panel where it stopped cleaning and hence continues with the cleaning process.

The PV power can be effectively used for charging the battery. The system can be readily installed for solar panels without any major modifications in the arrangement. The system is flexible and can be easily extended to meet the cleaning requirements of an entire solar farm. Further optimizations in size and cost can be achieved by using smaller circuit boards and building a smaller robot.

Using more powerful batteries, vacuum motors and gripper wheels would be helpful in implementing the robot on a larger scale. This work can be extended by implementing dust sensing methods to control the speed of the rolling brush and the vacuum motor. This system is very helpful to check if an area is cleaned thoroughly. If the robot is unable to clean a spot it can be programmed to send a message to the user so that the particular spot can be cleansed manually to avoid formation of hotspots. Addition of a gyroscope to the navigation control of this system would improve the feedback mechanism. Implementing machine learning concepts would enable the robot to find preset values by itself accurately, thus making it fully autonomous and a step closer to be released as a commercial product.

**Appendix A**

**PCB Designs**

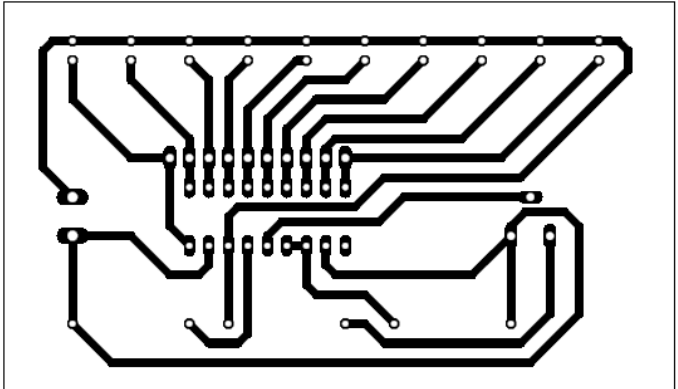

**Fig A.1.1: Battery Level Indicator**

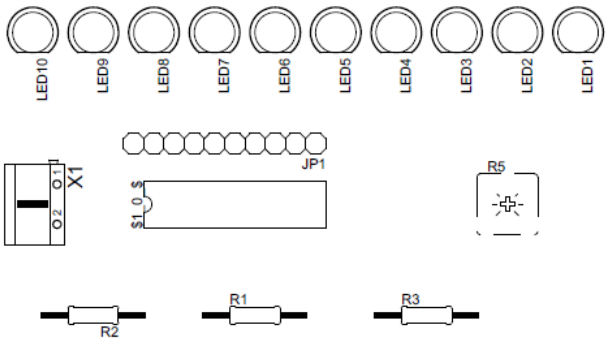

**Fig A.1.2: Battery Level Indicator**

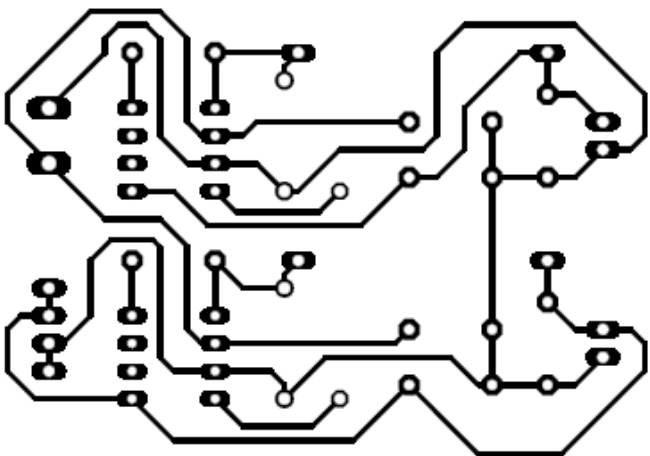

**Fig A.2.1: DC-DC Buck Converter**

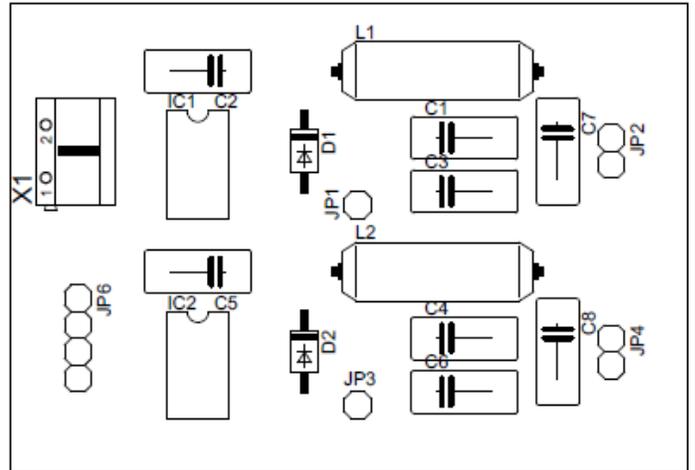

**Fig A.2.2: DC-DC Buck Converter**

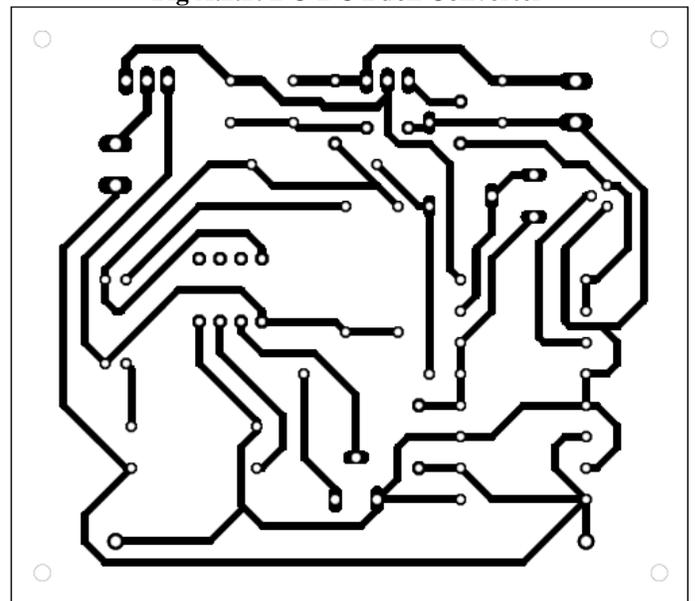

**Fig A.3.1: Lithium Polymer Battery Charging Circuit**

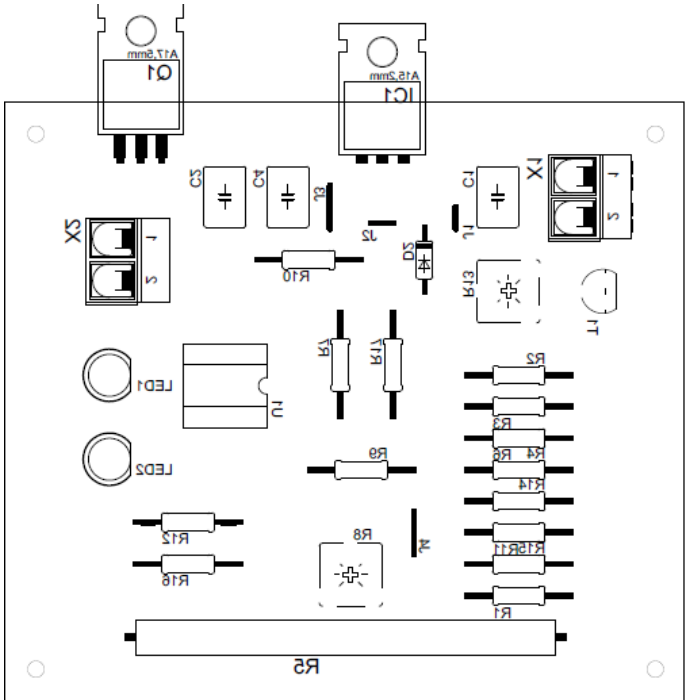
**Fig A.3.2 Lithium Polymer Battery Charging Circuit**

**Appendix B**

```c
#include <msp430.h>
#include <stdio.h>
#include <math.h>

// Pin Configuration for motors attached to wheels
#define right1 BIT4
#define right2 BIT3
#define left1 BIT1
#define left2 BIT2

// Configure Ultrasonic Sensor operation
#define trig BIT0
#define echo BIT5
#define battery_level BIT1
#define no_of_samples 5
#define cliff_threshold 4
#define turn_tolerance 5
#define ultrasonic_error_tolerance 20

/* Preset values used for accelerometer. Though the same x1
value can be used theoretically for all the 3 conditions, different
values are used since values change in practical testing
environments */
#define x1_set_up 218
#define x1_set_down 218
#define y1_set_turn 247
#define x1_set_turn 217

// PWM reference for speed control
#define ref_up 800
#define ref_down 100

// PID Constants
#define kp 20
#define ki 2
#define kd 10

// Definition of Global Variables
int x, ADC, turn_c=0, turn_k=0, c=0; distance_from_dock=0,
int battery_was_low=0;
char buffer[32],count=0;
float i,set,ref,x1=0.0,y1=0.0,z1=0.0,x2=0.0,y2=0.0,z2=0.0;
float error=0.0,ep=0.0,ei=0.0,ed=0.0,e=0.0,duration, distance;
static float prev_error = 0.0;

void trigger_ultrasonic();
int check_ultrasonic();

void initPWM()    //Initialize PWM
{
   P2DIR |= left2 + right2 + left1 + right1 ;
   P2SEL |= right1 +  left1;

   TA1CCR0 = 1000;
   TA1CCTL1 = OUTMOD_7;     TA1CCTL2 = OUTMOD_7;
   TA1CTL = TASSEL_2 + MC_1;
}

// Initialize ADC
void initADC()
{

ADC10CTL0 = SREF_0 + ADC10SHT_0 + ADC10ON;
// Use Vcc/Vss for Up/Low Refs, 16 x ADC10CLKs, turn on ADC

ADC10CTL1 = SHS_0 + ADC10SSEL_3 + ADC10DIV_1 +CONSEQ_0;
//  use ADC10CLK div 1, single channel mode

ADC10AE0 = 0xF9;
//  0,3,4,5,6,7 for ADC  from two accelerometer
}

unsigned int read_adc(int pin)
{
ADC10CTL0 = 0;
ADC10CTL1 = 0;
initADC();
switch(pin)
{
case 0: ADC10CTL1 |= INCH_0;break;
case 6: ADC10CTL1 |= INCH_6;break;
```

```c
case 7: ADC10CTL1 |= INCH_7;break;
case 3: ADC10CTL1 |= INCH_3;break;
case 4: ADC10CTL1 |= INCH_4;break;
case 5: ADC10CTL1 |= INCH_5;break;
}
ADC10CTL0 |= ENC + ADC10SC;
while ((ADC10CTL1 & ADC10BUSY) == 0x01);
// wait for conversion to end
return(ADC10MEM);
}

void stop(void)
{
TA1CCR1 = 1000;  TA1CCR2 = 1000;
P2OUT |=(left2+right2);}
void reverse_it(void)
{
TA1CCR1=200;TA1CCR2=200;P2OUT|=(left2+right2);
_delay_cycles(2500000);
}

void go(int a)
{
for(i=0;i<no_of_samples;i++)   {
x1=x1+read_adc(3)+read_adc(0);
y1=y1+read_adc(4)+read_adc(6);
z1=z1+read_adc(5)+read_adc(7);
}

x1/=2*no_of_samples;
y1/=2*no_of_samples;
z1/=2*no_of_samples;
x1=round(100*(((x1/1024.0)*3.3))/.8);
y1=round(100*(((y1/1024.0)*3.3))/.8);
z1=round(100*(((z1/1024.0)*3.3))/.8);
switch(a)
{
case 0:
set = x1_set_down; ref = ref_down;
error = x1 - set; break;
case 1:
set = y1_set_turn;ref = ref_down;
error = x1 - set;break;
case 2:
set = x1_set_down; ref = ref_down;
error = y1 - set;break;
}
ep = error; ei += error; ed = error - prev_error;
e = (ep*kp) + (ei*ki) + (ed*kd);
switch(a)
{
case 0:
if(ref+e<0) TA1CCR1 = 0; else TA1CCR1 = ref+e;
if(ref-e<0) TA1CCR2 = 0; else TA1CCR2 = ref-e; break;

case 1:
if(ref - e > 1000) TA1CCR1 = 1000;
else if(ref - e < 0 )TA1CCR1 = 0;
else TA1CCR1 = ref - e;
if(ref + e > 1000) TA1CCR2 = 1000;
else if(ref + e < 0 )TA1CCR2 = 0;
else TA1CCR2 = ref + e;
break;

case 2:
TA1CCR1 = ref - e;  TA1CCR2 = ref + e;
}
P2OUT &=~(left2+right2);
prev_error=error;
}

void turn(int x,int y,int a,int b)
{
for(i=0;i<no_of_samples;i++)   {
x1=x1+read_adc(3)+read_adc(0);
y1=y1+read_adc(4)+read_adc(6);
z1=z1+read_adc(5)+read_adc(7);
}
x1/=2*no_of_samples; y1/=2*no_of_samples;
z1/=2*no_of_samples;
x1=round(100*(((x1/1024.0)*3.3))/.8);
y1=round(100*(((y1/1024.0)*3.3))/.8);
z1=round(100*(((z1/1024.0)*3.3))/.8);
if(b==0)  error =  x1 - x1_set_turn;  //aligning by using X value
else  error =  y1 - y1_set_turn;        //aligning by using Y value
ep = error; ei += error; ed = error - prev_error;
e = (ep*kp) + (ei*ki) + (ed*kd);
if(a==1) e*=-1;
if(e>=0)                             {
TA1CCR1 = 500 - e; TA1CCR2 = 500 + e;
P2OUT &=~(left2);           P2OUT |= right2;
if(turn_c==1){ turn_c=0;turn_k++; }
                                       }
else{
TA1CCR1 = 500 - e; TA1CCR2 = 500 + e;
P2OUT |= (left2); P2OUT &=~ right2;
if(turn_c==0){ turn_c=1;turn_k++;}
      }
}

void trigger_ultrasonic()
{
P2OUT |= trig ; _delay_cycles(1000);
P2OUT &= ~(trig);
}

int check_ultrasonic()
{
int error=0;
duration=0.0; distance=0.0;
```

```c
while(error<40)
{
if(P2IN & echo){ duration++;}    else error++;
}
distance = (duration/2.0)/74.07;
if(distance*100 > cliff_threshold) return 0;
 //ultrasonic range cliff detect
else { return 1;}
}

void turn_it(int x, int y, int a, int b)
{
while(turn_k < turn_tolerance) turn(x,y,a,b); turn_k=0;
//turn_tolerance pid orientation alignment
stop(); _delay_cycles(10000);
}

void move_it(int x)
{
while(1)
{
trigger_ultrasonic();
if(check_ultrasonic()){ go(x); }
else { stop();
c++;
if(c > ultrasonic_error_tolerance ){ c=0; break; } }
_delay_cycles(10000);
//minimum time period for consecutive echoes from ultrasonic
if(x==2) distance_from_dock++;
    }
}

int side_move_it(int f)
{
if(f==0)
{
for(count=0;count<10;count++)
{
trigger_ultrasonic();
if(check_ultrasonic()){TA1CCR1=300;TA1CCR2=300;
P2OUT&=~(left2+right2); }
else {reverse_it();              break;}
_delay_cycles(100000);
}
}
else
{
for(count=0;count<10;count++)
{
trigger_ultrasonic();
if(check_ultrasonic()){
TA1CCR1=300;TA1CCR2=300;
P2OUT&=~(left2+right2);
}
else {
reverse_it();
turn_it(0,1,0,0);
turn_it(0,0,1,1);
return 1;
}
_delay_cycles(10000);
}
}
return 0;
}

void main(void)
{
   WDTCTL = WDTPW + WDTHOLD;
   P2DIR|=left1+right1+left2+right2;  P1REN |= battery_level;
   initADC();
   initPWM();
   while(1)
   {
    while(1)
    {
    x:move_it(1);
     reverse_it();
     turn_it(0,0,0,1);
     side_move_it(0);
     turn_it(0,1,1,0);
     move_it(0);
reverse_it(); reverse_it();
if(battery_level==0)
{turn_it(1,0,1,1);battery_was_low=1;break;}
turn_it(1,0,0,1);
if(side_move_it(1)) break;
turn_it(0,1,0,0);
}
move_it(2);
turn_it(1,1,0,0);
move_it(0);
stop();
if(battery_level == 1 && battery_was_low == 1)
{
reverse_it();turn_it(1,0,0,1);
while(distance_from_dock--)
go(2);
turn_it(0,1,0,0);
goto x;}
for(x=0;x<1000;x++)
 _delay_cycles(8640000);
}
}
```

**Appendix C**

| S.No | Components | Manufacturer | Cost per Component (₹) | Quantity | Total Cost (₹) | TI Supplied/Purchased |
|---|---|---|---|---|---|---|
| | MSP430G2553 Launchpad | Texas Instruments | 600 | 1 | 600 | Yes |
| | LM3914 | Texas Instruments | 60 | 1 | 60 | Yes |
| | LM317 | Texas Instruments | 30 | 1 | 30 | Yes |
| | TPIC0298 Motor Driver Module | Texas Instruments | 300 | 2 | 600 | No |
| | LM2675 - 5.0 | Texas Instruments | 216 | 1 | 216 | Yes |
| | LM2675 - 3.3 | Texas Instruments | 216 | 1 | 216 | Yes |
| | LM358N | Texas Instruments | 6 | 1 | 6 | Yes |
| | MMA7361 Triple Axis Accelerometer | Extreme Electronics | 400 | 2 | 800 | No |
| | HCSR04 - Ultrasonic Sensor and clamp | NA | 250 | 1 | 2 | No |
| | IRFZ44n | ST Microelectronics | 30 | 1 | 4199 | No |
| | IRF540n | ST Microelectronics | 35 | 1 | 600 | No |
| | BJT - BC547 | ST Microelectronics | 2 | 1 | 990 | No |
| | Lithium Polymer Battery | Zippy | 4199 | 1 | 200 | No |
| | High Speed Vacuum Motor | NA | 600 | 1 | 515 | No |
| | High Torque DC Geared Motors | Vega Robo Kits | 330 | 3 | 250 | No |
| | Wheels & Gears | NEX-Robotics | 200 | 2+3 | 30 | No |
| | Mechanix Kit | ZEPHYR | 515 | 1 | 35 | No |
| | L-Clamps & Screws | NEX-Robotics | 100 | - | 100 | No |
| | Wood, Acrylic Sheet, etc | NA | 700 | - | 700 | No |
| | Aluminium Strip | NA | 100 | - | 100 | No |
| | Rolling Brush | NA | 30 | - | 30 | No |
| | PCB Printing | NA | 300 | - | 300 | No |
| | Miscellaneous | NA | 250 | - | 250 | No |
| | TOTAL COST | | | | 10829 | |